\documentclass[11pt]{article}

\usepackage[final]{acl}

\usepackage{times}
\usepackage{latexsym}

\usepackage[T1]{fontenc}

\usepackage[utf8]{inputenc}

\usepackage{microtype}

\usepackage{inconsolata}

\usepackage{graphicx}

\usepackage{bookmark}   
\hypersetup{
	bookmarks=true,
	bookmarksopen=true,
	bookmarksnumbered=true,
	pdftitle={Mwando: Leveraging AI to Preserve and Teach shiKomori},
	pdfauthor={Naira Abdou Mohamed et al.},
}

\usepackage{tcolorbox}
\tcbuselibrary{skins, breakable}
\usepackage{ifthen}

\newtcolorbox{userbox}[2][]{
	colback=blue!5!white,
	colframe=blue!75!black,
	arc=4mm,
	boxrule=0.5pt,
	left=2mm,
	right=2mm,
	top=1mm,
	bottom=1mm,
	sharp corners=southwest,
	fontupper=\normalsize,
	before upper={\textbf{User: }},
	#1
}

\newtcolorbox{assistantbox}[2][]{
	colback=green!5!white,
	colframe=green!50!black,
	arc=4mm,
	boxrule=0.5pt,
	left=2mm,
	right=2mm,
	top=1mm,
	bottom=1mm,
	sharp corners=southeast,
	fontupper=\normalsize,
	before upper={\textbf{Assistant: }},
	#1
}

\newtcolorbox{analysisbox}[2][]{
	colback=gray!5!white,
	colframe=gray!30!white,
	arc=2mm,
	boxrule=0.3pt,
	left=2mm,
	right=2mm,
	top=1mm,
	bottom=1mm,
	fontupper=\itshape\small,
	before upper={\textbf{Analysis: }},
	#1
}

\usepackage{amsmath}

\usepackage{graphicx}

\usepackage{booktabs}

\title{\textit{Mwando}: Leveraging AI to Preserve and Teach shiKomori}

\author{
	Naira Abdou Mohamed$^{1}$\thanks{Contact: \texttt{naira@rifailabs.com}},
	Haidar Nassur Said Ali$^{2}$,
	Mohamed Hazra$^{3}$, \\
	\textbf{Naoufal Mohamed Soibira}$^{1,4}$,
	\textbf{Roushnaty Ali Yamani}$^{3}$\\
	$^1$ Rifai, Moroni, Comoros \\
	$^2$ CP2BM Lab - Hassan II University, Casablanca, Morocco \\
	$^3$ Ibn Tofail University, Kenitra, Morocco \\
	$^4$ Sciences Po Grenoble, Grenoble, France
}

\begin{document}
\maketitle
\begin{abstract}
This paper presents \textit{Mwando}, a virtual educational assistant designed to support the teaching and preservation of shiKomori, the language of the Comoros Islands. The system covers the four main dialectal variants (shiNgazidja, shiMwali, shiNdzuani and shiMaore) through a knowledge base constructed from phrases, proverbs, dictionaries and grammar lessons. A multi-agent architecture combining vector search, a knowledge graph and web search fallback enables accurate and context-aware responses. Evaluation on 500 queries demonstrates strong performance on vocabulary lookup and grammar explanations, while qualitative case studies illustrate both capabilities and current limitations. This work represents an initial step toward computational support for shiKomori and provides a blueprint for developing AI-powered educational tools for other low-resource languages.
\end{abstract}

\section{Introduction}
Language preservation is a central concern for international organizations such as the United Nations Development Programme (UNDP), which emphasizes African languages as key drivers of the continent development \cite{UNDPMIMIT2024LanguageData}. However, this potential remains largely unrealized, as very few technological solutions currently support these languages \cite{Wild2025}. This situation is problematic, as language and culture are major drivers of development \cite{Rotondo2016} and play a central role in everyday life across African societies. However, advances in language processing technologies, particularly generative AI systems, offer a promising opportunity: when appropriately adapted to cultural and linguistic specificities, these models can serve as effective conduits for the preservation of cultural heritage \cite{Colace2025,koc2025generativeailargelanguage}.

In this work, we aim to contribute to the initial development of Natural Language Processing (NLP) technologies for shiKomori, the language spoken in the Comoros Islands. Building on our previous work on foundational NLP resources for Comorian and its dialectal variations \cite{naira-etal-2024-datasets,naira-etal-2025-preserving}, we shift the focus to an applied educational setting.

Specifically, we introduce Mwando, a virtual assistant designed for teaching shiKomori, with the goal of supporting the transmission and preservation of Comorian culture. Mwando ("beginning" in shiKomori) honors Sheikh Ahmed Kamar-Eddine's pioneering work on Comorian language documentation and writing standardization \cite{lafon:halshs-00265704,naira-etal-2025-preserving}. Our contributions are threefold:


\begin{itemize}
	\item We compile and organize a multi-source corpus for shiKomori, covering phrases, proverbs, dictionaries and grammar lessons and make these data publicly available to support future research.
	\item We develop a virtual assistant for educational purposes, capable of interacting with learners in shiKomori.
	\item We evaluate the assistant’s potential for supporting language learning and cultural preservation, providing insights for further development of NLP resources for extremely low-resource languages.
\end{itemize}

\section{Motivations}
The arrival of generative AI has been a serious game changer in education \cite{learnlmteam2025evaluatinggeminiarenalearning} and it was immediately adopted by students and language teachers \cite{Zaim2025}. Among the reasons justifying this rapid adoption is the possibility for students to obtain quick feedback through a virtual teacher and for teachers, the ability to quickly and effectively customize lessons according to different cases \cite{galaczi_luckin_2024_genai_language_education}. It is precisely along these lines that a project was adopted in Senegal in elementary education to support French teachers in better teaching in classrooms composed of students with significant cultural and linguistic diversity \cite{afd_2025_ai_senegal}.

In the Comoros, although the use of AI and information technology in education is very poorly documented \cite{Roukiyat2026}, there are previous studies that have emphasized the importance of taking shiKomori into account in teaching for greater effectiveness \cite{DANIEL_2024}. And local initiatives such as the Maecha association have set themselves the goal of literacy in Comorian for the adult population, which is composed of approximately 49.7\% illiterate individuals \cite{Chauvet2015}.

Consequently, it is crucial to consider modern solutions for learning shiKomori. Such initiatives could not only benefit other categories of people and sectors, such as improving the experience of tourists in the archipelago, but they would also play an important role in the country's sustainable development \cite{Roukiyat2026}. Furthermore, they would hold strong potential for the inclusion of descendants of the Comorian diaspora, particularly those in France, whose population is estimated at more than 300,000 people \cite{lemonde_2024_comores_diaspora} and who contribute significantly to the development of the archipelago \cite{abdillahi:tel-01206102}.

\section{Linguistic Resources}
One of the biggest challenges for this research was obtaining the data. Our approach was to consult every possible resource available in order to build our knowledge base. However, to ensure the reliability of the dataset, manual checks were necessary at each stage of processing. In this section, we therefore describe all the corpora used, from the raw data to the transformed data for this work.

\subsection{Language Overview}
Spoken exclusively in the Comoros archipelago, shiKomori is a Bantu language, closely related to Swahili \cite{AhmedChamanga2022} and to the Sabaki language group \cite{serva2021}. It consists of four dialectal varieties, each primarily associated with one island, yet exhibiting a high degree of mutual intelligibility among speakers \cite{AhmedChamanga2022,naira-etal-2024-datasets}. ShiKomori holds the status of a national language in the archipelago, while French and Arabic are the official languages. In practice, shiKomori is mainly used orally in everyday communication and informal media, French dominates administrative and formal educational domains and Arabic is predominantly used in religious contexts \cite{Chauvet2015}.

In written form, shiKomori can be transcribed using either the Latin alphabet or the Arabic script \cite{lafon:halshs-00265704,naira-etal-2025-preserving}. However, its written use remains limited and is characterized by a lack of orthographic standardization. Despite recent initiatives aimed at improving the representation of African languages in NLP \cite{adebara-etal-2025-evaluating}, shiKomori remains largely underrepresented in the field, except for a few prior works \cite{naira-etal-2024-datasets,naira-etal-2025-preserving,abdourahamane:hal-01992871}.

\subsection{Data Sources}
\begin{figure*}
	\centering
	\includegraphics[width=0.8\linewidth]{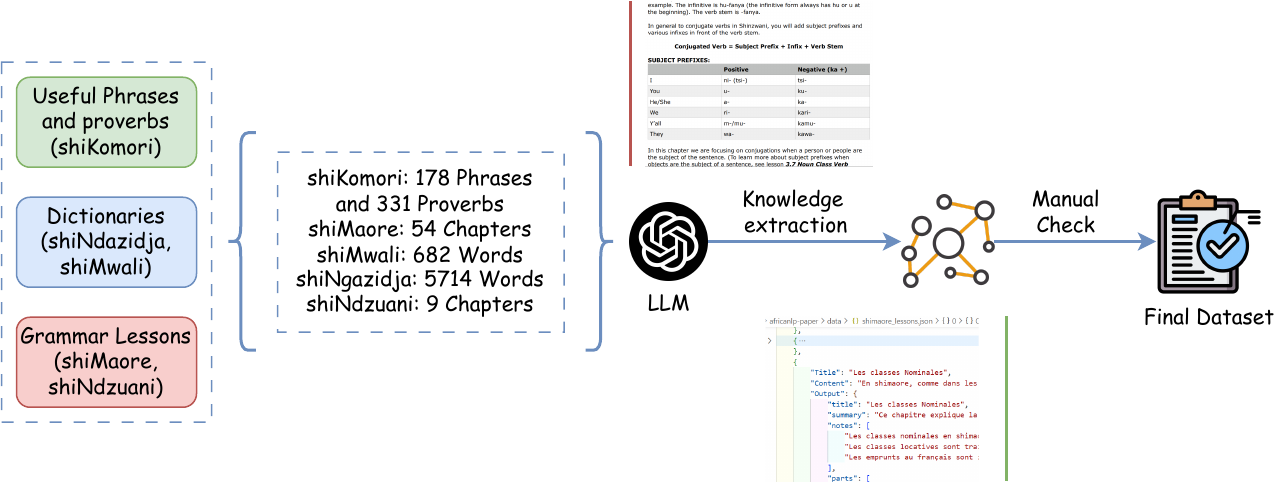}
	\caption{Data Preparation Pipeline.}
	\label{fig:data}
\end{figure*}

This work relies on six primary data sources, illustrated in Figure \ref{fig:data}. These sources are grouped into four main categories of data:
\begin{itemize}
	\item \textbf{Useful phrases}: These consist of commonly used phrases in shiKomori, without distinction between dialectal varieties. The phrases are translated into English and cover a wide range of topics, from simple greetings to expressions commonly used in contexts such as markets, travel and everyday interactions. The data were obtained from a Google Drive repository gathering various resources related to the Comoros Islands\footnote{\url{https://drive.google.com/drive/folders/17C\_03qCMm2rGDhgGqi\_8PKJfR30V\_S96}}.
	
	\item \textbf{Proverbs}: The proverbs were collected from the Instagram page \emph{ProverbesComoriens}\footnote{\url{https://www.instagram.com/proverbescomoriens/}} using the Apify scraping tool. Each post contains proverbs in shiKomori along with their French and English translations. However, the posts are published as images. To extract the textual content, we used DeepSeek-OCR \cite{wei2025deepseek}. This OCR-based extraction occasionally introduced minor errors, particularly with diacritics and special characters, which were manually corrected during the data cleaning phase.
	
	\item \textbf{Dictionaries}: Two dictionaries were used. The first, focusing on shiNgazidja, is a notable work produced by the Bahari Foundation\footnote{\url{https://fr.scribd.com/document/619345660/ShiNgazidja-English-Dictionary}}, which provides shiNgazidja words and expressions translated into English, with explicit annotation of nominal and grammatical classes. The second dictionary was obtained from the same Google Drive repository mentioned above and contains shiMwali words translated into English, along with grammatical class annotations.
	
	\item \textbf{Grammar lessons}: The grammar lessons are in shiMaore and shiNdzuani and were obtained from the platforms The Swiss Bay\footnote{\url{https://theswissbay.ch/pdf/Books/Linguistics/Mega\%20linguistics\%20pack/African/Niger-Congo/Bantu/Shimaore\%2C\%20Manuel\%20Grammatical\%20de.pdf}} and LiveLingua\footnote{\url{https://www.livelingua.com/peace-corps/Comorian/Shinzwani\%20Grammar\%20Book-\%201st\%20ed\%202015.pdf}}, respectively.
	
\end{itemize}

Finally, all data sources used in this work are publicly available online. For the Instagram data, we only collected publicly accessible posts and complied with the platform's terms of service. The dictionaries and grammar manuals are freely distributed by their respective publishers or hosting platforms. We release our processed dataset under an open license to facilitate future research on Comorian language processing (see Table \ref{tab:datasets}).

\begin{table*}
	\centering
	\small
	\caption{Overview of the datasets used in this work.}
	\label{tab:datasets}
	\begin{tabular}{@{}lllp{5.5cm}@{}}
		\toprule
		\textbf{Category} & \textbf{Dialect(s)} & \textbf{\# Rows} & \textbf{Link} \\
		\midrule
		Useful Phrases  & All varieties & 178   & \url{https://huggingface.co/datasets/rifailabs/shikomori_phrases} \\
		Proverbs    & All varieties & 329   & \url{https://huggingface.co/datasets/rifailabs/shikomori_proverbs} \\
		Dictionary     & shiNgazidja   & 5,710 & \url{https://huggingface.co/datasets/rifailabs/shingazidja_dictionary} \\
		Dictionary      & shiMwali      & 682   & \url{https://huggingface.co/datasets/rifailabs/shimwali_dictionary} \\
		Grammar Lessons    & shiMaore      & 54    & \url{https://huggingface.co/datasets/rifailabs/shimaore_lessons} \\
		Grammar Lessons    & shiNdzuani    & 67    & \url{https://huggingface.co/datasets/rifailabs/shindzuani_lessons} \\
		\bottomrule
	\end{tabular}
\end{table*}

\subsection{Knowledge Base Construction}
A processing step was necessary to better structure the data. Indeed, among the collected data, for example, in the case of grammar lessons, some data were in the form of tables, as illustrated in Figure \ref{fig:data}. Although Large Language Models (LLM) are effective at understanding textual data, they exhibit limitations when processing text mixed with tabular content \cite{liu-etal-2024-rethinking}. This issue primarily arises during the data tokenization phase. To mitigate this, we first applied a document structuring step by converting the documents into Markdown format using the Word2md platform\footnote{\url{https://word2md.com/}}.

Subsequently, the Markdown content was fed into the ChatGPT API to restructure the data into a more easily exploitable knowledge base. This process involved extracting elements such as chapter sections, covered concepts, tables, definitions and related content and organizing them into JSON files. The specific structuring strategy varied depending on the type of data. In all cases, a manual validation step was conducted to review and correct the outputs generated by ChatGPT.

\section{Chatbot Workflow}
\begin{figure*}[h]
	\centering
	\includegraphics[width=0.8\linewidth]{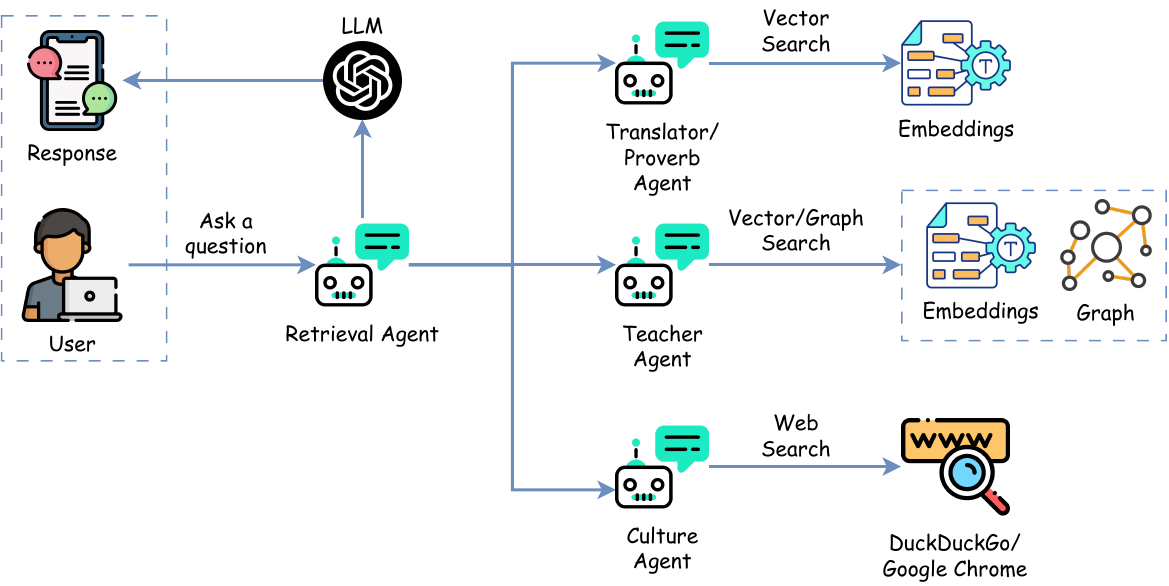}
	\caption{RAG Pipeline.}
	\label{fig:rag}
\end{figure*}

In recent years, the use of multi-agent approaches has significantly improved the reasoning capabilities of virtual assistants, particularly in situations where information must be retrieved from multiple sources \cite{salve2024collaborativemultiagentapproachretrievalaugmented}. In our context, where we need to handle multiple dialectal variants and types of datasets, adopting a multi-agent approach was a natural choice. Figure \ref{fig:rag} presents the overall pipeline that summarizes the core of our reasoning and information retrieval system.

To address the challenges of handling multiple Comorian dialects and heterogeneous data sources, we designed a multi-agent system that orchestrates specialized agents, each responsible for a specific type of query or knowledge domain. This architecture, illustrated in Figure \ref{fig:rag}, enables flexible and robust information retrieval by combining vector search, knowledge graphs and external web resources.

\subsection{Overall Pipeline}
The system operates as follows: when a user submits a query, it is first processed by a central Retrieval Agent. This agent acts as an orchestrator: it analyzes the query and determines which specialized agent should be invoked to provide the most accurate and comprehensive answer. The response is then generated by an LLM based on the retrieved information and returned to the user.

\subsection{Core Components and Specialized Agents}

The system comprises five main specialized agents, each leveraging different tools and knowledge bases:
\begin{itemize}
	\item \textbf{Retrieval Agent}: As the entry point for all queries, this agent is responsible for query understanding and dynamic routing. It decides whether to forward the request to a single specialist or to combine results from multiple agents. It relies on an LLM for intent classification and on a Vector Search engine to retrieve relevant passages from a precomputed embedding database covering all dialectal variants.
	\item \textbf{Translator/Proverb Agent}: This agent handles two closely related tasks. For translation queries, it accesses bilingual dictionaries (shiNgazidja-English, shiMwali-English) and phrase collections. For the queries related to proverb, we use a dedicated graph database that stores proverbs along with their meanings, cultural contexts and translations. The graph structure allows for semantic navigation (e.g., finding proverbs related to a specific theme).
	
	\item \textbf{Teacher Agent}: Focused on pedagogical needs, this agent provides grammar explanations, conjugation tables and language learning resources. It draws upon the grammar manuals (shiMaore and shiNdzuani) and can generate structured lessons or exercises. It also maintains a database of common learner errors and misconceptions.
	
	\item \textbf{Culture Agent}: For queries that fall outside the scope of other agents or require general information not covered by specialized knowledge bases, this agent performs live web searches. It serves as a fallback mechanism, ensuring that the system can still provide relevant answers even when internal resources are insufficient.
\end{itemize}

\subsection{Technical Implementation}
The system is built around several key technologies. For embeddings and vector search, all textual resources including dictionaries, proverbs, phrase collections and grammar lessons, were embedded using Qwen3 Embedding \cite{zhang2025qwen3embeddingadvancingtext}, a multilingual sentence transformer. This model was selected for two main reasons. First, it is among the top-performing embedding models on recent benchmarks, offering high-quality semantic representations across multiple languages. Second, with only 2 billion parameters, it is lightweight enough to run in resource-constrained environments, making it suitable for deployment in regions with limited computational infrastructure. The model was deployed locally using Ollama, a lightweight framework for running large language models efficiently on consumer hardware. The resulting embeddings were indexed in FAISS, a vector database \cite{douze2025faisslibrary} for efficient similarity search, enabling rapid retrieval of relevant passages based on query semantics rather than keyword matching alone.

Regarding the knowledge graph, proverbs and cultural knowledge are stored in a graph database using Neo4j. In this structure, nodes represent proverbs, definitions and everyday life entities, while edges capture semantic and thematic relationships. This design allows for nuanced navigation and reasoning across the knowledge base, which is particularly valuable for answering queries.

For orchestration and reasoning, agent coordination is implemented using LangChain. This handles prompt engineering, tool invocation, dynamic query routing and response synthesis. The core reasoning and language generation tasks are performed using Groq's ultra-fast inference platform, which hosts gpt-oss, an open-source LLM optimized for conversational AI. Additionally, the Culture Agent can perform live searches using DuckDuckGo when information is unavailable internally. Retrieved web content is summarized by gpt-osss before integration into the final response.

\section{Evaluation}

\subsection{Retrieval Performance}

For this evaluation, we set a target of testing 500 queries, divided among five evaluators with 100 queries each. Each evaluator categorized their queries according to the following four areas: Proverb Interpretation, Grammar Explanations, Vocabulary Lookup and Cultural Questions. We measured the following metrics, often used to evaluate the quality of a recommendation system \cite{Jadon2024} and more recently for evaluating a RAG system \cite{gan2025retrievalaugmentedgenerationevaluation}:

\begin{itemize}
	\item \textbf{Precision@k}: This metric measures the proportion of relevant documents among the top \textit{k} results.
	
	\item \textbf{Recall@k}: This metric measures the proportion of relevant items retrieved out of all existing relevant items. In other words, if there are \(p\) relevant items in total and among the top \(k\) results we find \(q\) relevant items, then Recall@k is \(q/p\).
	
	\item \textbf{Mean Reciprocal Rank (MRR)}: The average of the reciprocal ranks of the first relevant result. It is computed as follows:
	\[
	MRR = \frac{1}{N} \sum_{i=1}^{N} \frac{1}{\text{rank}_i}
	\]
	where \(N\) is the total number of queries and \(\text{rank}_i\) is the rank position of the first relevant result for query \(i\).
	
\end{itemize}

Table \ref{tab:retrieval-results} presents retrieval performance across different query categories. The system performs best on vocabulary lookup, which benefits from direct matches in dictionary entries. Grammar explanations and proverb interpretation yield lower scores due to the complexity of these queries and the variability in how concepts are expressed across grammar manuals. Cultural questions are the most challenging, as they often require synthesis from multiple sources or reliance on web search.

\begin{table*}
	\centering

	\caption{Retrieval performance by query category}
	\label{tab:retrieval-results}
	\begin{tabular}{@{}lccc@{}}
		\toprule
		\textbf{Query Category} & \textbf{Precision@5} & \textbf{Recall@5} & \textbf{MRR} \\
		\midrule
		Vocabulary Lookup        & 0.82                  & 0.76              & 0.81          \\
		Grammar Explanations     & 0.71                  & 0.60              & 0.73          \\
		Proverb Interpretation   & 0.69                  & 0.64              & 0.78          \\
		Cultural Questions       & 0.65                  & 0.63              & 0.69          \\
		\midrule
		\textbf{Average}         & \textbf{0.71}         & \textbf{0.65}     & \textbf{0.75} \\
		\bottomrule
	\end{tabular}
\end{table*}

\subsection{Response Quality}

The goal here is to evaluate the quality of the responses provided by the LLM. To do this, we revisited the 500 queries used during the retrieval evaluation to examine the responses generated by the LLM. These responses were evaluated using a 5-point Likert scale along the following three axes:

\begin{itemize}
	\item \textbf{Accuracy}: Is the information factually correct?
	\item \textbf{Completeness}: Does the response fully address the query?
	\item \textbf{Clarity}: Is the response easy to understand for a learner?
\end{itemize}

Table \ref{tab:human-evaluation} summarizes the average scores. The evaluators noted that vocabulary and grammar queries were generally well-handled, while proverb interpretations sometimes lacked the depth of cultural context that a human expert would provide. Cultural questions occasionally suffered from incomplete or overly general answers when relying on web search.

\begin{table*}
	\centering
	\caption{Human evaluation of response quality (average scores, 1-5 scale)}
	\label{tab:human-evaluation}
	\begin{tabular}{@{}lccc@{}}
		\toprule
		\textbf{Query Category} & \textbf{Accuracy} & \textbf{Completeness} & \textbf{Clarity} \\
		\midrule
		Vocabulary Lookup & 4.35 & 4.26 & 4.80 \\
		Grammar Explanations & 4.24 & 4.76 & 4.39 \\
		Proverb Interpretation & 4.08 & 3.82 & 4.15 \\
		Cultural Questions & 3.65 & 3.24 & 3.86 \\
		\midrule
		\textbf{Average} & \textbf{4.08} & \textbf{4.02} & \textbf{4.30} \\
		\bottomrule
	\end{tabular}
\end{table*}

\subsection{Case Studies}
\begin{figure*}[h]
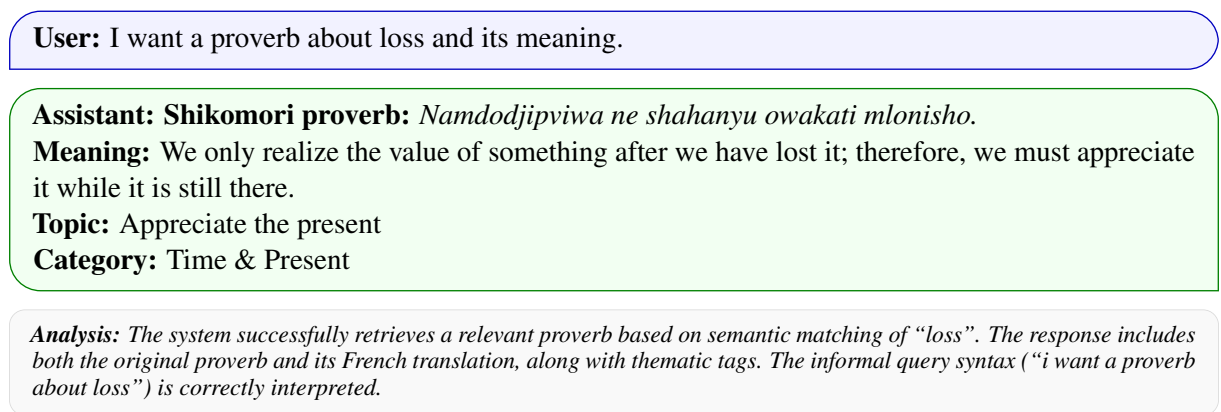

	\caption{Cultural Questions}
	
	\begin{userbox}{}
		What is the best way to greet an elderly person in ShiNdzuani?
	\end{userbox}
	
	\begin{assistantbox}{}
		The most appropriate greeting for an older person in ShiNdzuani is:
		
		\textbf{Mungu nahunusuru.}
		
		This is the customary salutation used when addressing someone older than yourself.
	\end{assistantbox}
	
	\begin{analysisbox}{}
		The system correctly identifies the culturally appropriate greeting for addressing elders in ShiNdzuani. This demonstrates its ability to capture social norms and respectful language use, which is essential for learners navigating Comorian cultural contexts. However, the response could be enriched with additional context, such as the literal meaning of the phrase (``May God protect you'') or the expected response from the elder.
	\end{analysisbox}
    \label{fig:exp3}
\end{figure*}

While the quantitative metrics presented in the previous section provide a high-level overview of system performance, they do not capture the nuances of individual interactions. To complement these findings, we present four representative case studies (See Figures \ref{fig:exp1}, \ref{fig:exp2}, \ref{fig:exp3} and \ref{fig:exp4}) that illustrate the system's behavior in practice. These examples span the four different query categories analyzed previously. Each case concludes with a brief discussion of strengths and limitations observed during the interaction.

\begin{figure*}[h]
	\caption{Vocabulary Query (shiNgazidja)}
	
	\begin{userbox}{}
		Translate the word ``hospital'' in shiNgazidja.
	\end{userbox}
	
	\begin{assistantbox}{}
		\textbf{Word:} lopitali / lapitali (malapitali) \\
		\textbf{Meaning:} ``hospital'' \\
		\textbf{Grammatical class:} noun (undefined) \\
		\textbf{Noun class:} 9‑10, 5‑6 \\
		\textbf{Origin:} French (fr)
	\end{assistantbox}
	
	\begin{analysisbox}{}
		The system correctly retrieves the dictionary entry with morphological variations (singular/plural) and etymology.
	\end{analysisbox}
	\label{fig:exp1}
\end{figure*}

\begin{figure*}[h]
	\caption{Proverb Interpretation}
	
	\begin{userbox}{}
		I want a proverb about loss and its meaning.
	\end{userbox}
	
	\begin{assistantbox}{}
		\textbf{Shikomori proverb:} \textit{Namdodjipviwa ne shahanyu owakati mlonisho.} \\
		\textbf{Meaning:} We only realize the value of something after we have lost it; therefore, we must appreciate it while it is still there. \\
		\textbf{Topic:} Appreciate the present \\
		\textbf{Category:} Time \& Present
	\end{assistantbox}
	
	\begin{analysisbox}{}
		The system successfully retrieves a relevant proverb based on semantic matching of ``loss''. The response includes both the original proverb and its French translation, along with thematic tags. The informal query syntax (``i want a proverb about loss'') is correctly interpreted.
	\end{analysisbox}
    \label{fig:exp2}

\end{figure*}

\begin{figure*}[h]
	\caption{Grammar Explanation (shiNdzuani)}
	
	\begin{userbox}{}
		How do noun classes work in shiNdzuani?
	\end{userbox}
	
	\begin{assistantbox}{}
		In shiNdzuani, every noun belongs to a \textbf{noun class}. The class is signalled by a \textbf{prefix} on the noun (and on the plural form of the noun). That same prefix is also used by adjectives, numbers, demonstratives and verbs so that they \textbf{agree} with the noun.
		
		\medskip
		\textbf{How the class is recognised:}
		\begin{itemize}
			\item Look at the singular prefix of the noun.
			\item The corresponding plural prefix tells you which class it is (the ``singular‑plural prefix correspondence'').
			\item Example: the singular prefix \textit{m-} (or \textit{mw-}) pairs with the plural prefix \textit{wa-} (this is class 1‑2).
		\end{itemize}

    	\medskip
        \textbf{Examples:}
        \begin{center}
            
            \begin{tabular}{llll}
                \toprule
                
                \textbf{Singular} & \textbf{Plural} & \textbf{Class} & \textbf{Meaning} \\
                \midrule
                gari     & magari  & class ? (m-/ma-) & ``voiture / voitures'' \\
                dzitso   & matso   & class ? (d-/ma-) & ``œil / yeux'' \\
                \bottomrule
            \end{tabular}
            
        \end{center}

		\medskip
		\textbf{Agreement:}
		\begin{itemize}
			\item An adjective placed after the noun must take the same class prefix: \textit{gari kubwa} (big car). If the noun were plural, the adjective would also change: \textit{magari makubwa}.
			\item Numbers and ordinal numbers behave the same way; most change form to match the class, except a few invariable ones (e.g., \textit{muhimu}, \textit{sita}, \textit{saba}, \textit{shendra}, \textit{kumi}).
			\item Verbs also carry the noun‑class prefix so that the subject and verb agree in class.
		\end{itemize}
		
		\medskip
		So, to use shiNdzuani correctly you:
		\begin{enumerate}
			\item Identify the noun's class by its singular‑plural prefix pair.
			\item Apply the same prefix to any modifying adjective, number, demonstrative, or verb.
			\item If a word is listed without a prefix change (e.g., the invariable adjectives above), it does not change with the noun class.
		\end{enumerate}
	\end{assistantbox}
	
	\begin{analysisbox}{}
		This response demonstrates the system's ability to provide structured grammatical explanations. It correctly identifies noun class mechanisms and provides clear examples with agreement rules.
	\end{analysisbox}
    \label{fig:exp4}
\end{figure*}

\subsection{Discussion of Limitations}

Our evaluation reveals several limitations:

\begin{itemize}
	\item \textbf{Coverage gaps}: Despite our efforts, some dialectal variants and specialized vocabulary remain underrepresented, particularly for shiNdzuani and shiMwali.
	\item \textbf{Cultural depth}: Proverb interpretations and cultural explanations sometimes lack the nuanced understanding that a human expert would provide.
	\item \textbf{Web search quality}: When relying on web search, the system occasionally retrieves low-quality or irrelevant information, affecting response accuracy.
	\item \textbf{Evaluation scale}: Our human evaluation involved only five annotators; a larger study with more diverse participants (learners, teachers, elders) would provide more robust insights.
\end{itemize}

Despite these limitations, the results demonstrate that \textit{Mwando} provides a solid foundation for computer-assisted language learning for shiKomori, with particular strength in vocabulary and grammar support.

\section{Conclusion and Future Work}
In this paper, we introduced \textit{Mwando}, a virtual educational assistant for shiKomori, supporting its four dialectal variants. We assembled a multi-source corpus comprising phrases, proverbs, dictionaries and grammar lessons and developed a multi-agent architecture combining vector search, a knowledge graph and web search fallback. The system leverages lightweight embeddings (Qwen3 via Ollama) and fast reasoning (Groq with gpt-oss, an open-source LLM).

Evaluation on 500 queries showed strong performance on vocabulary lookup and grammar explanations, with qualitative case studies illustrating both capabilities and current limitations, particularly in cultural depth and proverb interpretation.

Coverage remains uneven across dialects and web search quality is variable. Future work will focus on expanding resources for shiNdzuani and shiMwali, enriching cultural knowledge with expert input, improving web search filtering and deploying \textit{Mwando} in real-world educational settings. We hope this work contributes to the preservation of Comorian heritage and serves as a blueprint for AI support in other low-resource languages.

\bibliography{custom}

\end{document}